\documentclass[times,twocolumn,final,authoryear]{elsarticle}

\usepackage{prletters}
\usepackage{framed,multirow}

\usepackage{amssymb}
\usepackage{latexsym}

\usepackage{url}
\usepackage{xcolor}

\usepackage{graphicx}
\usepackage{caption}
\usepackage{subcaption}

\definecolor{newcolor}{rgb}{.8,.349,.1}

\journal{Pattern Recognition Letters}

\begin{document}
\thispagestyle{empty}
\begin{frontmatter}

\title{Open Turbulent Image Set (OTIS)}

\author[1]{J\'er\^ome Gilles\corref{cor1}} 
\cortext[cor1]{Corresponding author: 
  Tel.: +1-619-594-7240;  
  fax: +1-619-594-6746;}
\ead{jgilles@mail.sdsu.edu}
\author[1]{Nicholas B. Ferrante}

\address[1]{San Diego State University - Department of Mathematics \& Statistics, 5500 Campanile Dr, San Diego, CA 92182, USA}

\received{1 May 2013}
\finalform{10 May 2013}
\accepted{13 May 2013}
\availableonline{15 May 2013}
\communicated{S. Sarkar}

\begin{abstract}
Long distance imaging is subject to the impact of the turbulent atmosphere. This results into geometric distortions and some blur effect in the observed frames. Despite the 
existence of several turbulence mitigation algorithms in the literature, no common dataset exists to objectively evaluate their efficiency. In this paper, we describe a new 
dataset 
called OTIS (Open Turbulent Images Set) which contains several sequences (either static or dynamic) acquired through the turbulent atmosphere. For almost all sequences, 
we provide the corresponding groundtruth in order to make the comparison between algorithms easier. We also discuss possible metrics to perform such comparisons.
\end{abstract}


\end{frontmatter}


\section{Introduction}
\label{sec1}
Turbulence mitigation algorithms aiming to restore a clean image from a set of distorted observations has been widely studied since more than a decade. See for instance 
\cite{Frakes2001,Gilles2008,Mao2012c,Lou2013,Anantrasirichai2013,Halder2013a,Micheli2013,Gal2014} just to cite a few. Most contributions are based on the common idea of combining 
stabilization and deconvolution steps and they only differ on the techniques used in their implementation. Most of the time, each author has access to data which are not freely 
available hence making the comparison difficult. It becomes necessary to build an open dataset of images as well as select some metric which can be used by the community in order 
to get an objective comparison of the different developed algorithms.\\
The purpose of this paper is to propose such open dataset. It is made of a collection of sequences of static scenes as well as dynamic scenes (i.e with a moving target). Most of 
the static sequences correspond to the observation of printed charts. Such approach permits to create a groundtruth associated to each sequence and then can be used by some 
metric to assess the reconstruction efficiency.\\
We want to emphasize that the purpose of this dataset is not to assess the turbulence itself but the visual enhancements provided by mitigation algorithms.  Therefore, we do not provide any physical measurements (temperature, wind, $C_n^2$,\ldots) and we categorized the observed turbulence as either ``weak'', ``medium'' or ``strong''.\\
The paper is organized as follows. Section~\ref{sec:equip} presents the equipment used and the acquisition procedure. The acquired sequences are described in 
section~\ref{sec:data}. Possible evaluation metrics are discussed in section~\ref{sec:metric} and section~\ref{sec:conc} concludes this paper.

\section{Equipment and acquisition procedure}\label{sec:equip}

\subsection{Equipment}
\begin{figure}[!t]
\centering
\includegraphics[scale=0.08]{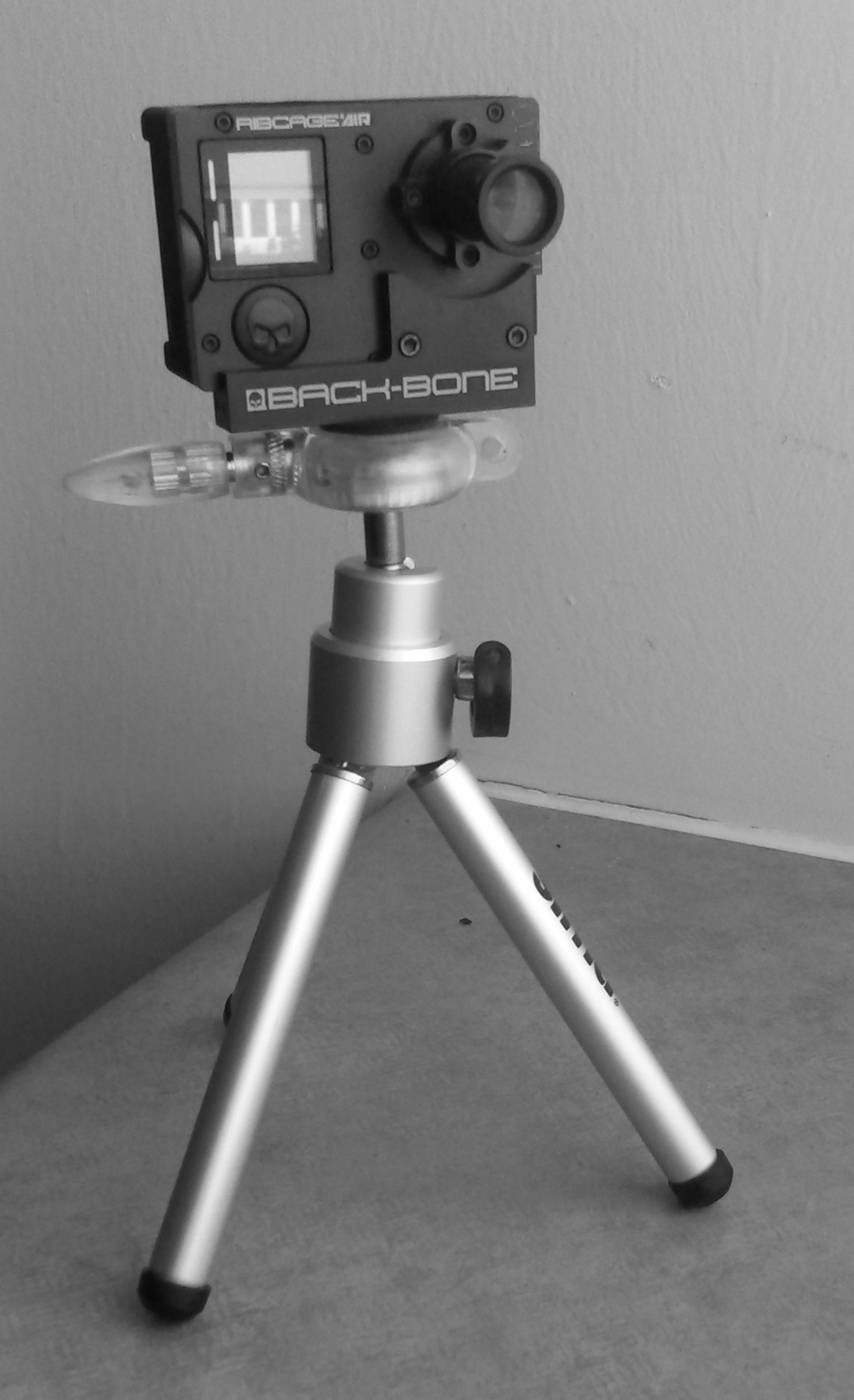}
\caption{RibCage Air Modified GoPro Hero 4 Black camera with a 25mm f/2.0 14d HFOV 3MP lens on its tripod.}
\label{fig:cam}    
\end{figure}

\begin{figure}[!t]
\includegraphics[width=0.49\columnwidth]{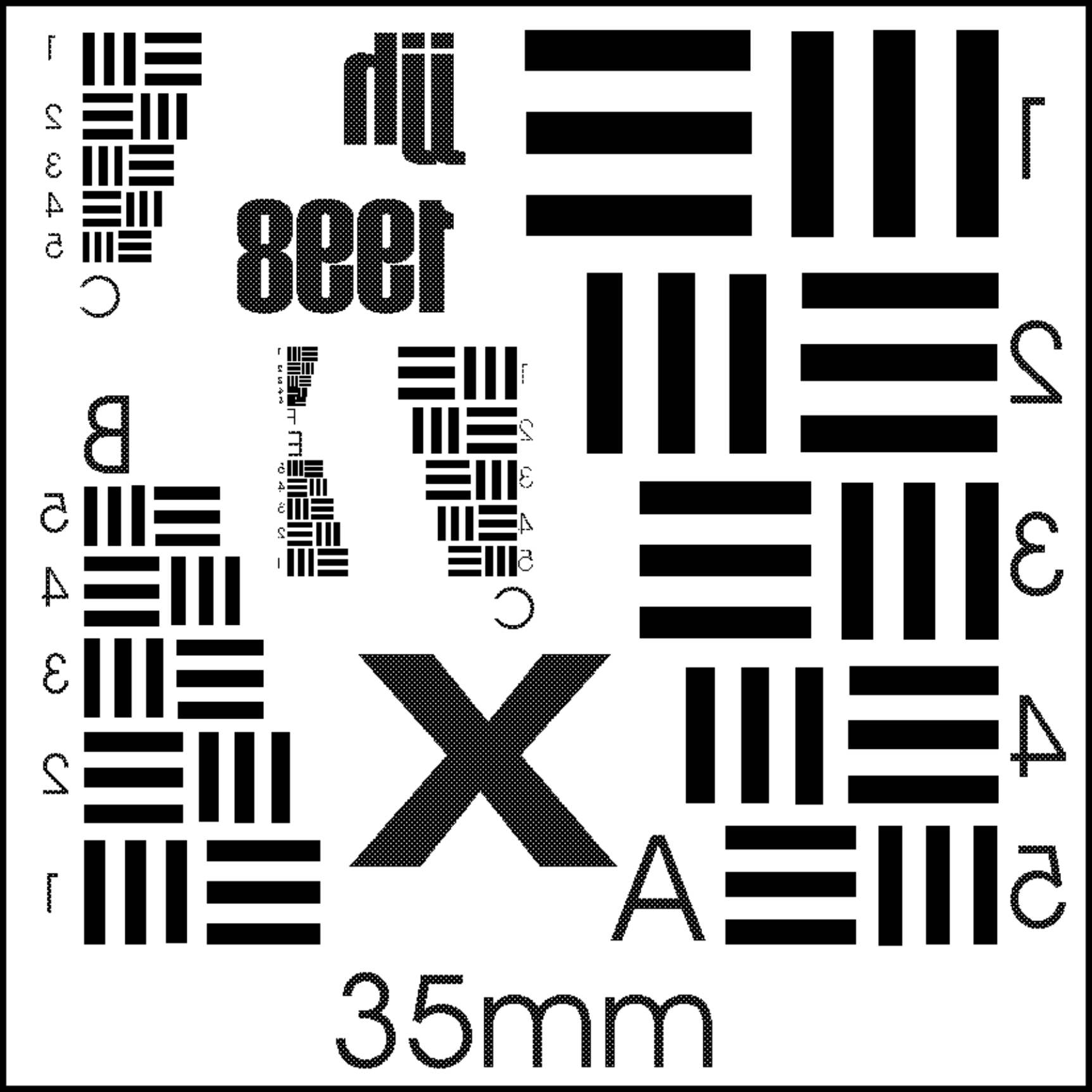}
\includegraphics[width=0.49\columnwidth]{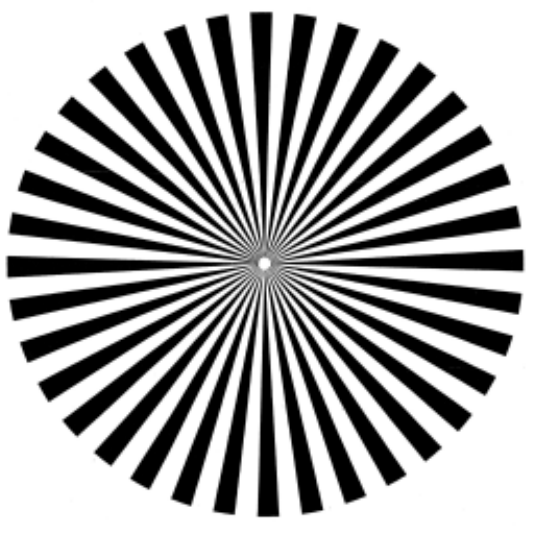}
\caption{The two charts serving as our static artificial targets after being printed on a poster.}
\label{fig:barchart}
\end{figure}

\begin{figure}[!t]
\centering
\includegraphics[width=\columnwidth]{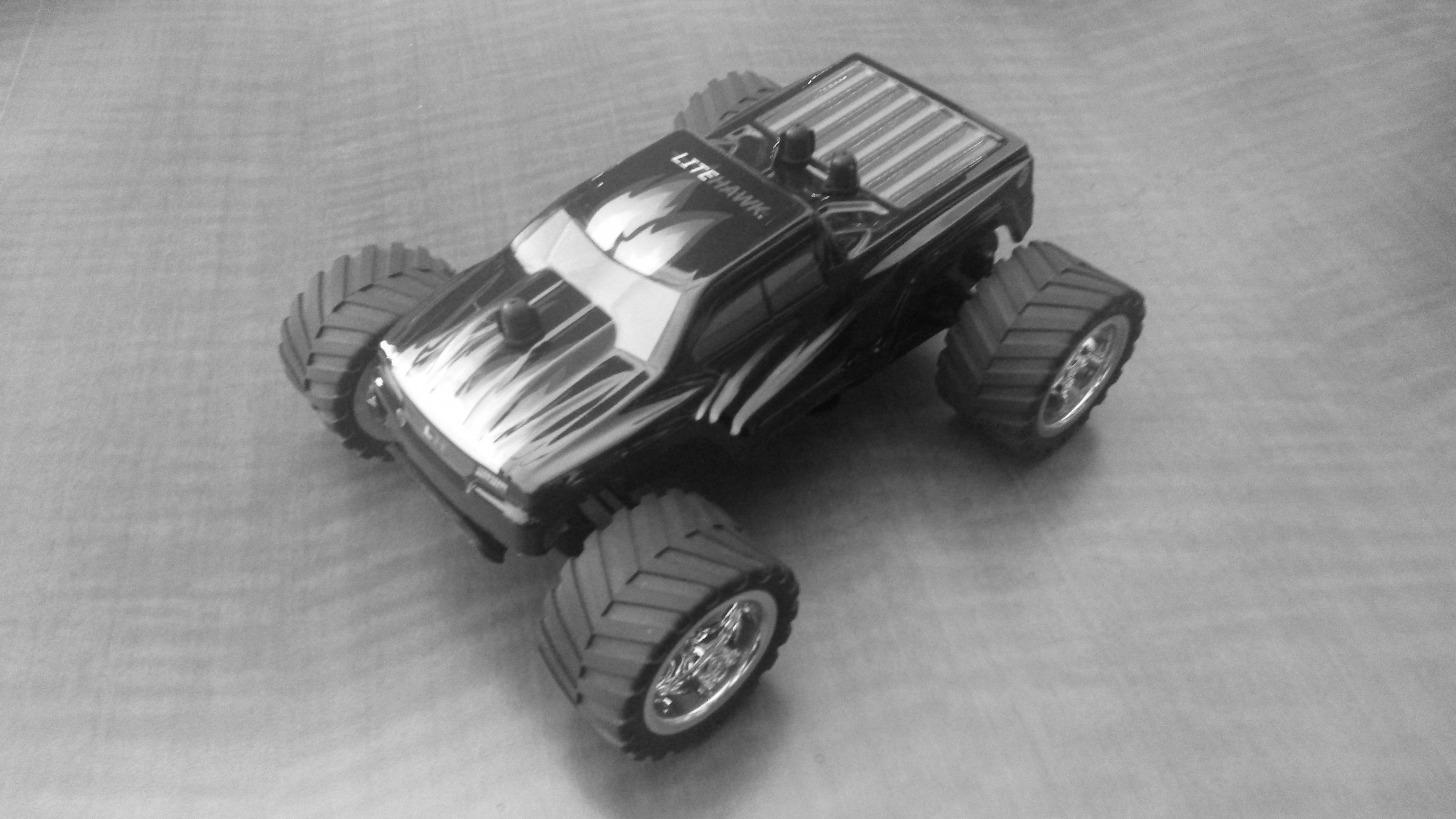}
\caption{Remote Controlled car utilized as our moving target for the dynamic sequences.}
\label{fig:car}
\end{figure}

All sequences were acquired with a GoPro Hero 4 Black camera modified with a RibCage Air chassis permitting to adapt several type of lenses. We always used a 25mm, f/2.0 14d HFOV 
3MP lens. The camera was setup at a 1080p resolution and a framerate of 24 frames per second (fps). A small tripod was used to hold the camera (see Figure~\ref{fig:cam}). The 
camera was controlled by the standard GoPro App on a Samsung Galaxy tablet.\\
The acquired sequences contain both natural elements from the observed scene as well as artificial ``targets''. For the static sequences, we used two charts containing some 
geometric patterns at different spatial frequencies and orientations (see Figure~\ref{fig:barchart}). These charts were printed on a poster (each chart has a size of 35x35cm) and 
held up by a homemade wooden stand. For the dynamic sequences, we used a standard remote controlled car (see Figure~\ref{fig:car}).

\subsection{Procedures}
\begin{figure}[!t]
\includegraphics[width=\columnwidth]{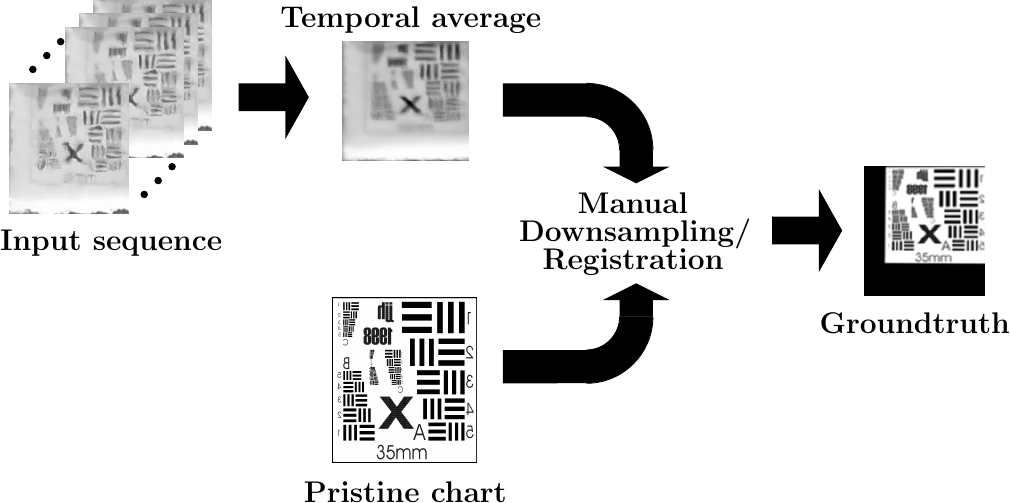}
\caption{Groundtruth images creation procedure.}
\label{fig:gtproc}
\end{figure}

All acquisitions were made on hot sunny days in order to guaranty a certain amount of atmospheric turbulence. All equipments were setup on a practice field at the San Diego State 
University. This field is equipped with artificial turf which reflects very well the sun heat, leading to high level of turbulence. The camera stood at about 
10cm above the ground observing the target positioned at several distances.\\
After all acquisitions were done, the different recorded movies were downloaded on a Linux computer and split into sequences of PNG image files using the \textit{ffmpeg} 
command\footnote{\url{https://ffmpeg.org/}}. The different region of interest are finally cropped via the \textit{convert} command (from the \textit{imagemagick} 
library\footnote{\url{http://www.imagemagick.org/}}) and saved as individual PNG sequences. Since the {Matlab\textregistered}  software is widely used by the community, we also 
provide each sequence saved as a Matlab 3D matrix (the first two coordinates are the spatial coordinates while the third one corresponds to time) save in a .mat file.\\
Since the purpose of this dataset is to be used for evaluating turbulence mitigation algorithms, all sequences containing the two above mentioned charts are provided with a 
groundtruth image. This groundtruth image contains the pristine chart after being downsampled and registered to the actual sequence. In practice, we manually registered the 
pristine 
chart on a temporal average of the input sequence using the GIMP\footnote{\url{https://www.gimp.org/}} software (this procedure is summarized in Figure~\ref{fig:gtproc}).\\
The dynamic sequences are also provided with their respective groundtruths. Each groundtruth is a sequence of binary images containing the bounding box corresponding 
to the moving target position. These groundtruths were created with the software 
\textit{Sensarea}\footnote{\url{http://www.gipsa-lab.grenoble-inp.fr/~pascal.bertolino/bin/download.php?file=sensarea.exe}}.

\section{Collected data}\label{sec:data}
The different available sequences were acquired between June 18th and August 16th, 2016.
\subsection{Static sequences}
\begin{figure}[!t]
\begin{tabular}{cc}
\includegraphics[width=0.45\columnwidth]{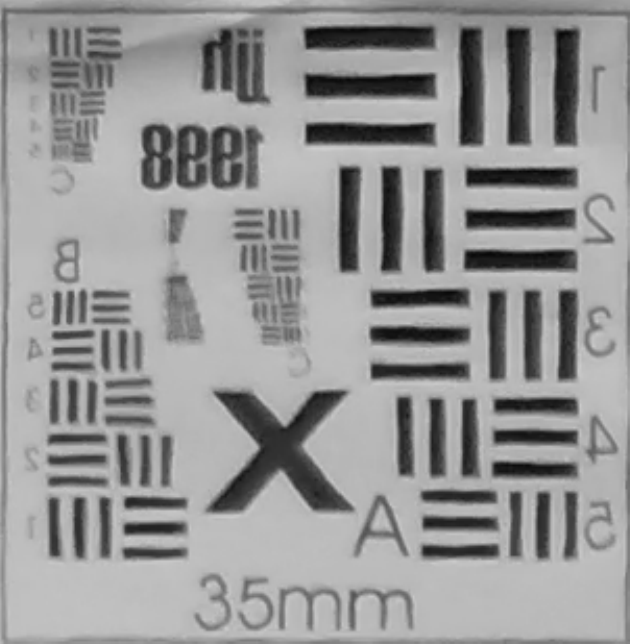} & \includegraphics[width=0.45\columnwidth]{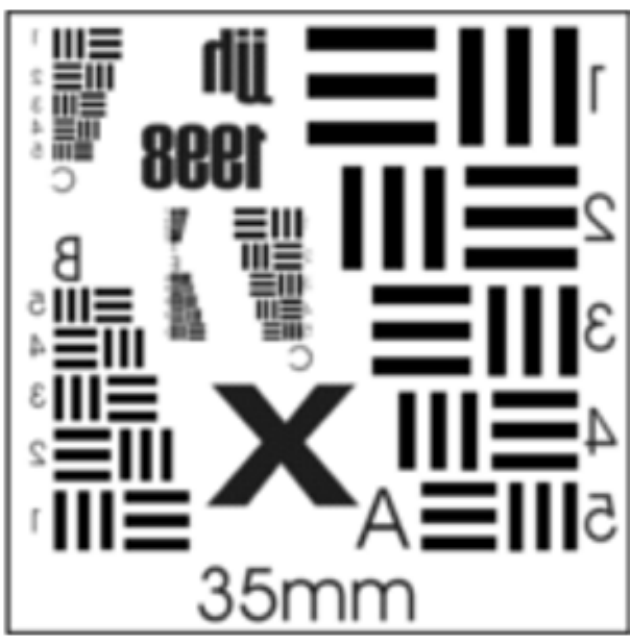} \\
\includegraphics[width=0.45\columnwidth]{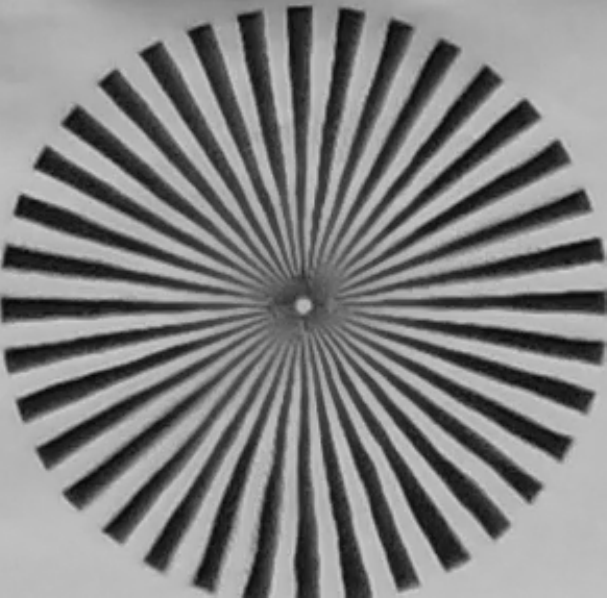} & \includegraphics[width=0.45\columnwidth]{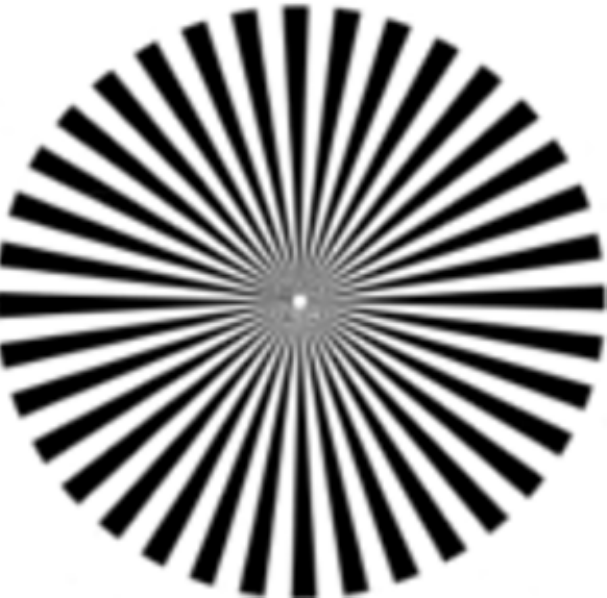} \\
\includegraphics[width=0.45\columnwidth]{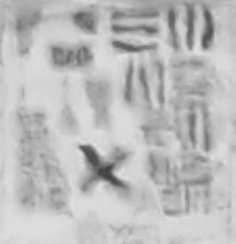} & \includegraphics[width=0.45\columnwidth]{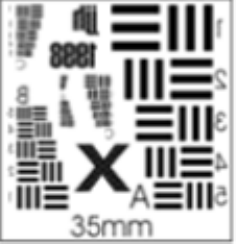} \\
\includegraphics[width=0.45\columnwidth]{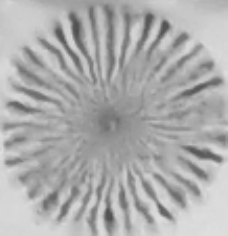} & \includegraphics[width=0.45\columnwidth]{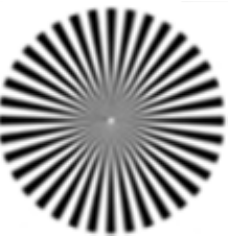}
\end{tabular}
\caption{Samples of frames from different fixed pattern sequences (left) and their corresponding groundtruths (right). The first two rows correspond to a weak turbulence case 
while the last two ones correspond to a strong turbulence case.}
\label{fixedpatterns}
\end{figure}

OTIS contains a total of seventeen static sequences. One provides the observation of natural elements in the scene (doors, steps, fence) and the sixteen remaining ones are made 
of the two previous charts. The former ones are provided with their corresponding groundtruth images in order to facilitate future algorithm evaluations. The complete list of all 
static sequences as well as their characteristics is given in Table~\ref{tab:static}. Several samples from different fixed pattern sequences as well as with their corresponding 
groundtruths are illustrated in Figure~\ref{fixedpatterns}. 
\begin{table*}[!t]
\caption{\label{tab1}Summary of the different static sequences}
\label{tab:static}
\begin{center}
\begin{tabular}{|p{3cm}|p{2cm}|p{1.5cm}|p{2cm}|p{2cm}|p{2cm}|} \hline
\textbf{Folder Name} & \textbf{Sequence Name} & \textbf{Number of images} & \textbf{Image size} & \textbf{Turbulence level} & \textbf{Ground Truth} \\  \hline
Fixed Background & Door & 300 & 520x520 & Strong & No \\  \hline
Fixed Patterns & Pattern1 & 64 & 302x309 & Weak & Yes \\ 
& Pattern2 & 64 & 291x287 & Weak & Yes\\ 
& Pattern3 & 300 & 113x117 & Strong & Yes\\ 
& Pattern4 & 300 & 109x113 & Strong & Yes\\ 
& Pattern5 & 300 & 113x117 & Strong & Yes\\
& Pattern6 & 300 & 109x113 & Strong & Yes\\
& Pattern7 & 300 & 122x125 & Strong & Yes\\
& Pattern8 & 300 & 119x122 & Strong & Yes\\
& Pattern9 & 300 & 152x157 & Medium & Yes\\
& Pattern10 & 300 & 149x149 & Medium & Yes\\
& Pattern11 & 300 & 172x183 & Medium & Yes\\
& Pattern12 & 300 & 172x173 & Medium & Yes\\
& Pattern13 & 300 & 202x202 & Weak & Yes\\
& Pattern14 & 300 & 196x193 & Weak & Yes\\
& Pattern15 & 300 & 134x139 & Strong & Yes\\
& Pattern16 & 300 & 135x135 & Strong & Yes\\  \hline
\end{tabular}
\end{center}
\end{table*}
\begin{figure}[!t]
\begin{tabular}{cc}
\includegraphics[width=0.45\columnwidth]{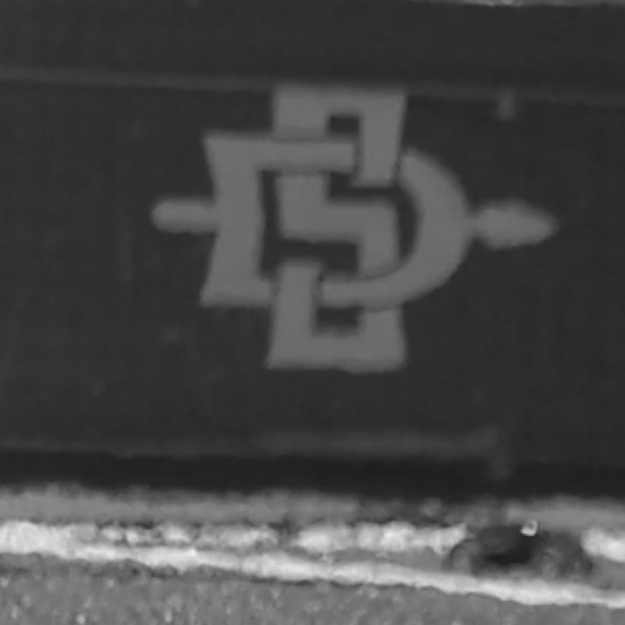} & \includegraphics[width=0.45\columnwidth]{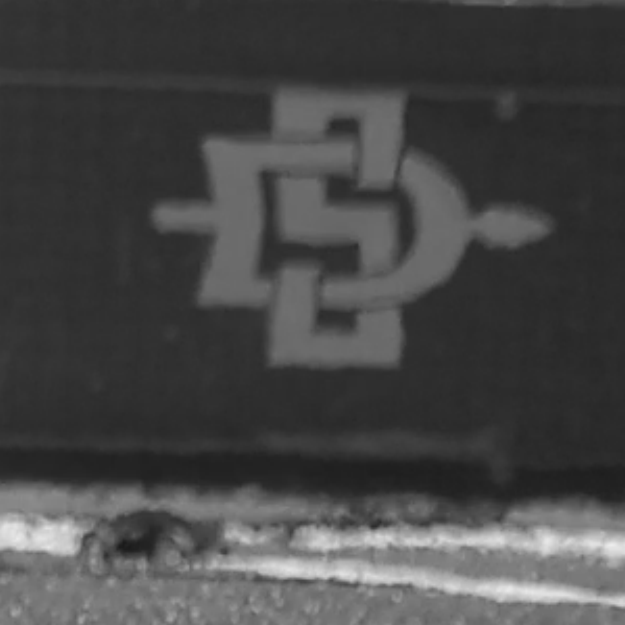} \\
\includegraphics[width=0.45\columnwidth]{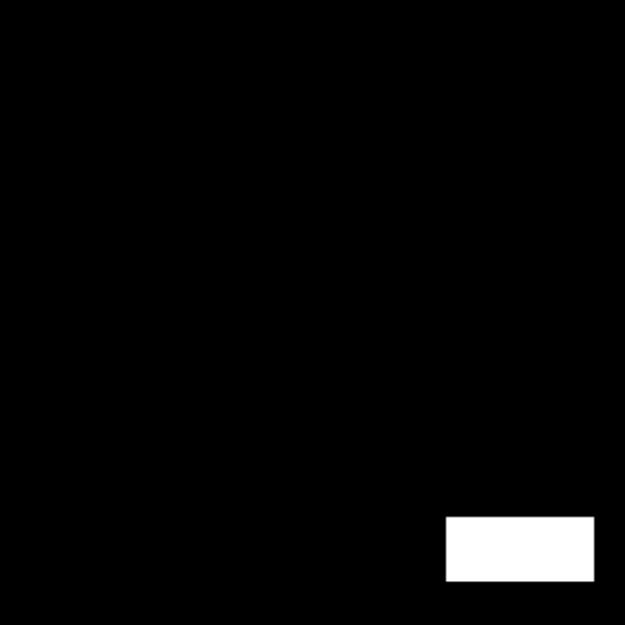} & \includegraphics[width=0.45\columnwidth]{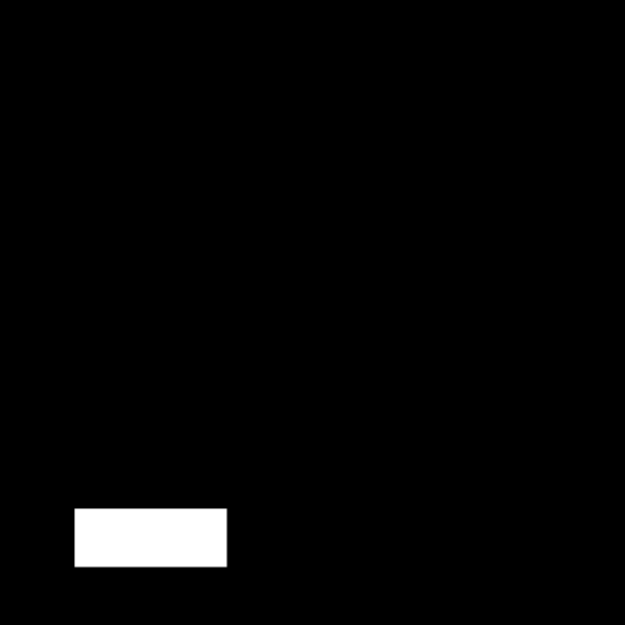} \\
\end{tabular}
\caption{Frames 16 and 34 from a dynamic sequences (top) and their corresponding groundtruths frames (bottom).}
\label{moving}
\end{figure}

\subsection{Dynamic sequences}
OTIS contains four dynamic sequences. Each sequence contains a moving remote controlled car at different distances and following different trajectories. Our goal was to 
create both sequences where the moving car remains easy to detect (even if it is affected by geometric distortions), as well as sequences where the car movement magnitude is close 
to the magnitude of the turbulence hence making them more challenging to detect. The complete list of all available sequences as well as their characteristics are given in 
Table~\ref{tab:dyn}. Two frames from a dynamic sequence with their corresponding bounding boxes are given in Figure~\ref{moving}.
\begin{table*}[!t]
\caption{\label{tab1}Summary of the different dynamic sequences}
\label{tab:dyn}
\begin{center}
\begin{tabular}{|p{3cm}|p{2cm}|p{1.5cm}|p{2cm}|p{2cm}|p{2cm}|} \hline
\textbf{Folder Name} & \textbf{Sequence Name} & \textbf{Number of images} & \textbf{Image size} & \textbf{Turbulence level} & \textbf{Ground Truth} \\  \hline
Moving Target & Car1 & 100 & 200x200 & Medium & Yes\\
& Car2 & 315 & 500x200 & Medium	& Yes\\
& Car3 & 51 & 300x300 & Medium & Yes\\
& Car4 & 101 & 300x300 & Medium & Yes \\  \hline
\end{tabular}
\end{center}
\end{table*}

\section{Quality metric}\label{sec:metric}
The main intent of OTIS is to facilitate the comparison of turbulence mitigation algorithms. Having test sequences as well as their groundtruth images is not sufficient to design 
a complete evaluation process.  Indeed, it remains to choose some metrics in order to obtain an objective comparison. Denoting $I_R$ and $I_{GT}$ the restored and groundtruth 
images, respectively, the most used metric to compare images is the Mean Square Error (MSE) defined by
$$MSE(I_R,I_{GT})=\frac{1}{N}\sum_x(I_R(x)-I_{GT}(x))^2,$$
where $N$ is the number of pixel in the image. Despite its popularity, the MSE has several drawbacks, notably it doesn't take into account the geometric information contained 
within the image. The Structural Similarity Index Measure (SSIM) was proposed by \cite{Wang2004} to circumvent these issues. The SSIM is a perception-based model which consider 
structural (geometrical) distortions in the image. Given the fact that one of the major degradation due to the turbulence is the geometrical distortions, the SSIM metric appears 
to be the most adapted metric to compare turbulence mitigation algorithms.\\
Regarding the assessment of tracking algorithms applied on the dynamic sequences, the methodology developed in \cite{Gilles2010} can be used.

\section{Conclusion}\label{sec:conc}
In this paper, we described a new publicly available dataset called OTIS (Open Turbulent Image Set) of sequences acquired through atmospheric turbulence which purpose is 
to aid at the assessment of mitigation algorithms. This dataset contains both static and dynamic sequences with groundtruth images. We propose OTIS in several versions: in color 
or grayscales and either saved as a set of PNG images or as 3D Matlab matrices. These different versions of OTIS can be downloaded for free at 
\url{https://zenodo.org/communities/otis/}. We also suggest the use of the SSIM metric to perform such objective performance evaluations.

\section*{Acknowledgments}
This work was supported by the NSF grant DMS-1556480, the San Diego State University Presidential Leadership Funds and the Air Force Office of Scientific Research grant FA9550-15-1-0065.

\bibliographystyle{model2-names}
\bibliography{refs}
\end{document}